# Sentiment Analysis Using Simplified Long Short-term Memory Recurrent Neural Networks *

Karthik Gopalakrishnan and Fathi M. Salem

*Department of Electrical and Computer Engineering*
*Michigan State University*
*East Lansing, Michigan-48824, USA*

{gopala17 & salemf}@msu.edu

*Abstract* – LSTM or Long Short Term Memory Networks is a specific type of Recurrent Neural Network (RNN) that is very effective in dealing with long sequence data and learning long term dependencies. In this work, we perform sentiment analysis on a GOP Debate Twitter dataset. To speed up training and reduce the computational cost and time, six different parameter reduced slim versions of the LSTM model (slim LSTM) are proposed. We evaluate two of these models on the dataset. The performance of these two LSTM models along with the standard LSTM model is compared. The effect of Bidirectional LSTM Layers is also studied. The work also consists of a study to choose the best architecture, apart from establishing the best set of hyper parameters for different LSTM Models.

## I. INTRODUCTION

Recurrent Neural Networks though in theory are capable of handling long-term dependencies fall short when it comes to practical applications. This problem was very well explored in depth by Hochreiter (1991) and Bengio, et al. (1994) [1]. It was seen that remembering information over long periods requires calculating the distances between distant nodes that involves multiple multiplications of the Jacobian Matrix. Problems with the more commonly occurring vanishing gradients and lesser frequent exploding gradients caused the performance of these models to be not satisfactory. It was seen that a trade of between gradient descent based learning and the time over which the information is held was required. In order to overcome this, Hochreiter and Schmidhuber (1997) [2] introduced the Long Short Term Memory networks usually called LSTM's. The LSTM's accumulates long-term relationships between distant nodes by designing weight coefficients between connections. These networks have shown unbelievable applications in speech processing, Natural Language Processing and image captioning among other applications.

A LSTM unit utilizes a "memory" cell (denoted by $c_t$) that decides whether the 'information' is useful or not and a gating mechanism that contains three non-linear gates: (i) an input (denoted by $i_t$), (ii) an output (denoted by $o_t$) and (iii) a forget gate (denoted by $f_t$). The gates regulate the flow of signals into and out of the cell to adjust long-term dependencies effectively and achieve successful RNN training. The standard equations for LSTM memory blocks are given as follows:

The three gates:

$$i_t = \sigma(U_i h_{t-1} + W_i x_t + b_i) \quad (1)$$
$$f_t = \sigma(U_f h_{t-1} + W_f x_t + b_f) \quad (2)$$
$$o_t = \sigma(U h_{t-1} + W x_t + b_o) \quad (3)$$

Memory Cells and hidden units:

$$c_t = i_t * c_{t-1} + \tanh(U_c h_{t-1} + W_c x_t + b_c) \quad (4)$$
$$h_t = o_t * \tanh(c_t) \quad (5)$$

Where $x_t$ is the external input vector, *tanh* is the hyperbolic tangent function, and the parameters are the matrices W and U, and vector bias b, with appropriate sizes for compatibility.
The output layer:

$$y_t = V * h_t + d \quad (6)$$

These parameters are all updated at each training step and stored. Though LSTM's or Recurrent Neural Networks in general take into consideration the past data, it is important to consider the future data for better performance. Bidirectional Recurrent Neural Networks (BRNN) [3] brings the future aspect into our network. BRNN's use two separate hidden layers that are then fed forward to the same output layer. The basic idea of BRNN is that each training sequence is represented by both in forward and backward directions respectively by two recurrent neural networks (RNNs), and that both are connected to one output layer. This structure provides complete past and future context information for each point in the output layer input sequence. This when combined with the LSTM's have shown promising results in dealing with the sequence processing problems such as continuous speech recognition [4], speech synthesis [5] and named entity recognition [6]. Figure 1 gives a basic architecture of a Bidirectional Recurrent Neural Network.

The internal dynamic "state" captures the essential information of an input's time-history profile. In any time series signal processing scheme in recurrent systems, repeated multiplication of internal states beyond the external input sequence isn't required or can be avoided. A matrix multiplication signifies scaling and rotation (which is basically mixing) of elements of a signal. This can also be achieved by scaling the subsequent processing after the matrix multiplication of the input signal. Scaling is generally expressed as a point wise (Hadamard) multiplication. The two observations above, once recognized can be exploited to define new families of Slim architectures of the LSTMs [9].

In this paper, a sentiment analysis on a publicly available dataset is done. Standard LSTM network is used as the baseline model and the performance of some of the reduced Slim LSTM models is compared. The positioning of the LSTM layer in the architecture is studied. A comprehensive set of hyper parameters is established as part of this work. Also, the effect of introducing bidirectional LSTM layers on the performance is studied. This paper is organised in five sections. Section II describes the relevant work done in this field, while section III goes over the Slim LSTM Models. Section IV covers all the experiments conducted as part of this study. The paper is concluded in Section V.

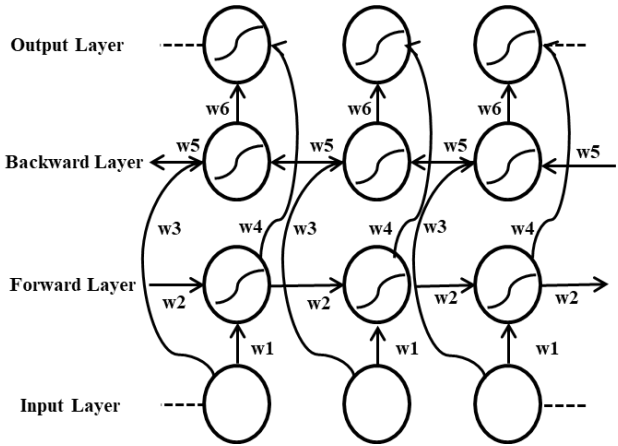

Figure 1: Basic Architecture of a BRNN

## II. LITERATURE REVIEW

Chung, et al examine and evaluate three different recurrent neural networks that basically are: LSTM-RNN, GRU-RNN and tanhRNN for the task of sequence modelling on two different datasets: polyphonic music data and raw speech signal data.[7] The results show that the convergence of gating units (GRU-RNN and LSTM-RNN) is much faster and their final solutions tend to be much better when compared with the traditional tanh-RNN on both the datasets. In order to compare the performance of LSTM and GRUs more concretely, Jozefowicz, et al examine the performance of LSTM and GRU on other datasets and also establish the other three best architectures discovered by the search procedure (named MUT1, MUT2, and MUT3).[8] They also evaluate an LSTM without input gates (LSTM-i), an LSTM without output gates (LSTM-o), and an LSTM without forget gates (LSTM-f). The main conclusions of these experiments are that: 1) GRU outperformed the LSTM on all tasks with exception of language modelling. 2) The LSTM with a large forget bias outperformed both LSTM and the GRU on almost all tasks. 3) That the LSTM-i and LSTM-o achieved the best results on the music dataset when dropout is used. Therefore, the LSTM-RNN and its derived structure proved to be the optimal and the best option to deal with the sequence model.

## III. SIMPLIFIED LSTM MODE

In order to reduce the adaptive parameter numbers in each gate, F. M. Salem [9] proposed six simplifications to the standard LSTM resulting in six LSTM variants by removing some of the parameters in the selected blocks and we refer to them here as simply, LSTM1, LSTM2, LSTM3, LSTM4, LSTM5 and LSTM6. In this way the computational cost can be reduced. The mathematical formulations of these six variants are given below.

A. LSTM1 MODEL: No input signal

$$i_t = \sigma(U_i h_{t-1} + b_i) \quad (7)$$
$$f_t = \sigma(U_f h_{t-1} + b_f) \quad (8)$$
$$o_t = \sigma(U h_{t-1} + b_o) \quad (9)$$

B. LSTM2 MODEL: No input signal, no bias

$$i_t = \sigma(U_i h_{t-1}) \quad (10)$$
$$f_t = \sigma(U_f h_{t-1}) \quad (11)$$
$$o_t = \sigma(U h_{t-1}) \quad (12)$$

C. LSTM3 MODEL: No input signal, no hidden unit signal

$$i_t = \sigma(b_i) \quad (13)$$
$$f_t = \sigma(b_f) \quad (14)$$
$$o_t = \sigma(b_o) \quad (15)$$

D. LSTM4 MODEL: No input signal, no bias & pointwise multiplication

$$i_t = \sigma(U_i \odot h_{t-1}) \quad (16)$$
$$f_t = \sigma(U_f \odot h_{t-1}) \quad (17)$$
$$o_t = \sigma(U \odot h_{t-1}) \quad (18)$$

E. LSTM5 MODEL: No input signal, no pointwise Multiplication

$$i_t = \sigma(U_i \odot h_{t-1}) + b_i \quad (19)$$
$$f_t = \sigma(U_f \odot h_{t-1}) + b_f \quad (20)$$
$$o_t = \sigma(U \odot h_{t-1}) + b_0 \quad (21)$$

F. LSTM6 MODEL: No input gate, No forget gate, bias is between (-1, 1)

$$i_t = 1.0 \quad (22)$$
$$f_t = \alpha, \ -1 < |\alpha| < 1 \ (default=0.59) \quad (23)$$
$$o_t = 1.0 \quad (24)$$

The first three models have been demonstrated previously in initial experiments in [10] using the MNIST and IMDB

dataset. In this work, we detail and try to demonstrate the comparative performance of the standard LSTM (baseline) and two Slim LSTM Models (LSTM 1 and LSTM6).The texts in this dataset have been already divided into negative sentiments (denoted by 0) and positive sentiments (denoted by 1), and the neural sentiment texts are deleted because it becomes hard to have words that acts as identifiers to a neutral sentiment which causes the performance of the model go down. Analysing Neutral Sentiments is planned to be presented in future research.

## IV. EXPERIMENTS

### A. Dataset

The dataset comes from Crowdflower from the "everyone library" [11] which contains tens of thousands of tweets about the early August GOP debate in Ohio for the 2016 presidential nomination. Each of them has been accordingly annotated with a sentiment label: positive, negative or neural. As mentioned, we neglect the neutral sentiments in the dataset.

For data pre-processing, the following steps were taken:

1) Selecting data: There are three types of sentiments in this dataset: the positive, the negative and the neutral sentiments. To make the models more efficient, we choose the positive and negative sentiments as training samples for reasons as mentioned before. Neutral sentiments generally don't have any fixed patterns and has more unstable factors and this makes it difficult to extract its features when compared to the other two classes. Also if positive and negative sentiments are classified accurately, we can filter out the neutral sentiments naturally.

2) Text pre-processing: All the words were converted into a lower format, and the characters were restricted to only letters and numbers, thus restricting the amount of generated features. If we consider more features, we would need more complex networks to analyse them.

3) Tokenizing: In order to have optimal performance, all the sentences were turned into number sequences. Each sequence is then padded with 0 to unify the dimensions of these sequences.

A count of the distribution of the data is shown in Figure 2 and it is observed that the data is extremely unbalanced; the number of negative samples was significantly more compared to the positive samples. This directly has an effect on the performance of the system as the network classifies negative sentiments way more accurately compared to positive samples. Artificially balancing the training dataset is certainly useful in classification problems, but there is always a chance of losing data, and sometimes important features of the training dataset. Since there are about almost more than 2000 samples in the minority class (Positive), the training dataset is not artificially balanced in this work, and is used for training as it is.

### B. Neural Network Structure and Parameters

The baseline architecture used contains 11 layers. At the top of the network is the embedding layer that is used to decrease the training dataset from 20000 dimensions to 128 dimensions of the dense embedding by transforming words to their corresponding word embedding's.

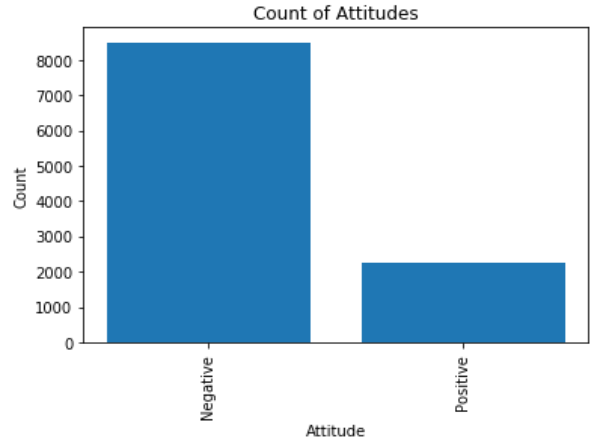

Figure 2: Dataset Split

As the words in the sequence are relevant with each other, the spatial dropout 1D layer is used to improve the independence between them. Then a 1D convolution layer and a max-pooling layer are used to help extract the features. Three different LSTM cells, which are: the original LSTM, LSTM1 (with no input) and LSTM6 (with no input and forget gates) are used, followed by a bidirectional LSTM model with the sigmoid activation to improve the performance of the network in processing the sequential data. At last, the three consecutive dense layers with dropout are used to change the dimensions of the output vector. The architecture demonstrated above is the overall structure for our LSTM neural network. The number of parameters varies according to different models. There are lesser number of parameters to train for the slim LSTM variants.

### C. Performance of the baseline LSTM0, LSTM1 and LSTM6 Models

The training accuracy and losses are plotted for 10 epochs. The testing accuracies are similar to that of the training values but suffer from a small amount of overfitting. Figure 3 shows the performance of the simple LSTM Model, while figure 4 and 5 show the performance of the simplified slim LSTM Models LSTM1 and LSTM6 respectively.

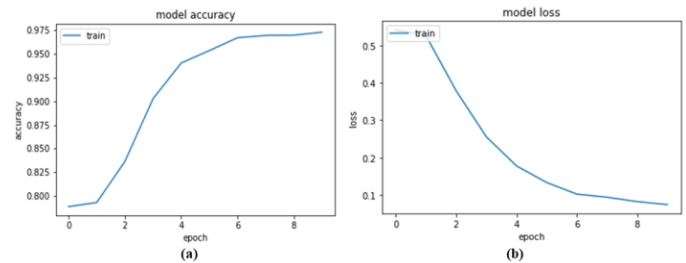

Figure 3: (a) Training Accuracy of the standard LSTM (Baseline) Model (b) Training Loss of the standard LSTM (Baseline) Model

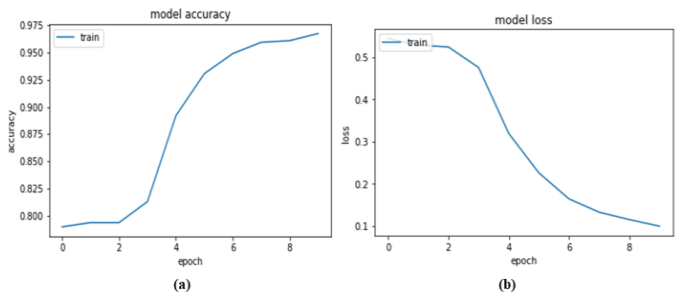

Figure 4: (a) Training Accuracy of the LSTM1 Model (b) Training Loss of the LSTM1 Model.

Table 1 meanwhile gives the testing accuracy for the three models, where all the models have been trained using learning rate of 0.0001 and an Adam optimizer. It is seen that baseline LSTM and the simplified LSTM6 model performs best. This is promising as a slimmed down version of the LSTM model can perform as well as a normal LSTM model. Therefore, similar results can be achieved by a slimmed down version that is computationally more efficient.

D. Effect of variations in Architecture on Performance

i) Positioning of LSTM Layer

The placement of LSTM layer was studied here. The baseline was a LSTM Layer and a 1D Convolutional Layer following it. In this paper, we have tried extracting the features through the 1D Convolutional Layer first and then passing it through the LSTM Layer and any other LSTM variant.

TABLE 1: TEST ACCURACY OF THE LSTM MODELS

| MODEL | TEST ACCURACY |
|---|---|
| LSTM0 | 84 |
| LSTM1 | 80 |
| LSTM6 | 83 |

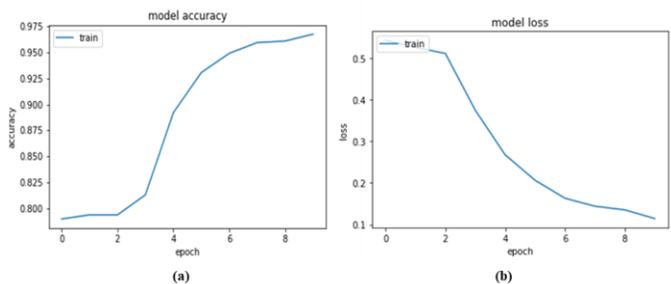

Figure 5: (a) Training Accuracy of the LSTM6 Model (b) Training Loss of the LSTM6 Model.

The rest of the architecture and the parameters are left unchanged. CNN-LSTM denotes a CNN Layer followed by a LSTM layer while LSTM-CNN denotes a LSTM Layer followed by a CNN Layer. Figure 6 details the performance of the two variants. It is observed that LSTM-CNN block performs very poorly in classifying positive sentiments and the negative sentiment accuracy is comparable for both the variants. Therefore, as expected, having a CNN layer initially to extract features, and then passing the extracted features to the LSTM layer yields in a better performance overall as it can classify both positive and negative sentiments accurately.

ii). Effect of multiple dense layers on performance

To the baseline architecture, three additional dense layers along with varying dropout rates were added after the LSTM block. Though it did increase the overall number of parameters and the training time, it was seen that adding dense layers helped improve the performances of standard LSTM (LSTM0) and the LSTM1 slightly but not so for the LSTM6 model. Table 2 gives the accuracies of the three models before and after adding dense layers.

D. Optimizers, Learning Rates and Batch Size

Learning rates along with batch sizes play a very big role in the performance of a network or a model. They are often termed as hyper-parameters and fine tuning them to optimal values is one of the challenges a researcher working in this field faces.

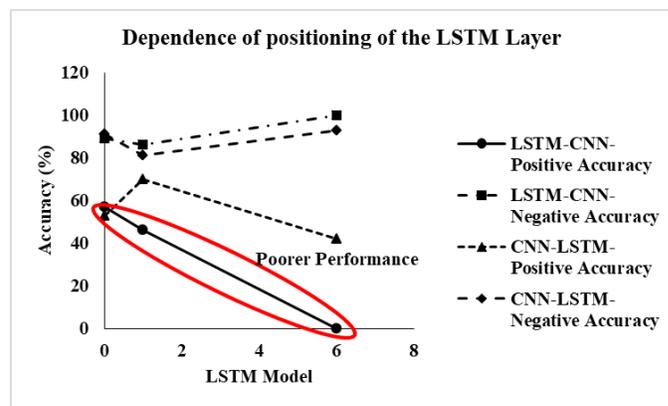

Figure 6: Exploring the positioning of the LSTM Layer

TABLE 2: EFFECT OF EXTRA DENSE LAYERS

| Model | Positive (%) | Negative (%) | Overall[1] | Overall[2] |
|---|---|---|---|---|
| LSTM0 | 57 | 90 | 83 | 84 |
| LSTM1 | 45 | 94 | 84 | 80 |
| LSTM6 | 62 | 86 | 81 | 83 |

**1**: After Adding  **2**: Before Adding

There is no single rate that can be used; rather it depends on the architecture, the dataset and the other hyper-parameters. In this work, we study the difference in performances when we use two different learning rates, two different optimization functions and four different batch sizes. Table 3 gives the performance of the network when run over 10 epochs with two different learning rates. From the table, the overall performance is the same for both the learning rates, but when it comes to classifying positive and negative sentiments individually, it can be concluded that a learning rate of 0.001 works best for our models. Table 4 meanwhile gives the performance of the system when exposed to the RMSprop activation. From the above table, we observe that RMSprop improves performance of LSTM1 considerably but not LSTM0 and LSTM6 though the performances of the LSTM6 model in both cases are comparable.

TABLE 3: EFFECT OF LEARNING RATE

|       | Positive | Negative | Positive | Negative | Overall |
|-------|----------|----------|----------|----------|---------|
|       | lr=0.001 |          | lr=3e-4  |          |         |
| LSTM0 | 53       | 91       | 58       | 88       | 82      |
| LSTM1 | 70       | 81       | 64       | 86       | 82      |
| LSTM6 | 42       | 93       | 60       | 88       | 82      |

It is therefore very hard to narrow down on an optimizer for this dataset, but on the whole RMSprop looks to work the best for the networks when we consider the positive and negative sentiment classification. The performance is comparable as Adam is a combination of the RMSprop and Stochastic Gradient Descent (SGD) with momentum. It utilizes the squared gradients to adaptively scale the learning rate like RMSprop as well as the moving average of the gradient (instead of the gradient itself) like SGD with momentum [12].

TABLE 4: EFFECT OF DIFFERENT OPTIMIZERS

| Model | Adam     |          |         | Rmsprop  |          |         |
|-------|----------|----------|---------|----------|----------|---------|
|       | Positive | Negative | Overall | Positive | Negative | Overall |
| LSTM0 | 53       | 91       | 84      | 67       | 83       | 80      |
| LSTM1 | 70       | 81       | 80      | 40       | 96       | 84      |
| LSTM6 | 42       | 93       | 83      | 71       | 81       | 79      |

Table 5 gives the performance of the networks when run over batch sizes 16,32,64 and 128. As expected the best performance is when the batch size is 16. This is so because smaller the batch size, more samples it trains in an epoch as it runs more iterations through the training dataset. This also results in longer training duration which is undesirable. We can also observe that the performance of all the models is almost similar.

E. Training Data Size

We also alter the training data size by tuning the training-validation data split ratio i.e if the ratio is increased there are lesser samples for training and more samples for testing or validation and vice versa. Table 6 gives the performance of the network when the ratio is 0.33 and 0.4. This goes against common intuition that is more the number of training samples, better the learning, and better the performance. This isn't true always, as the networks when subjected to similar data multiple times can over fit and underperform.

TABLE 5: EFFECT OF BATCH SIZE

| Model | BS=16 |     | BS=32 |     | BS=64 |     | BS=128 |     | OA* |
|-------|-------|-----|-------|-----|-------|-----|--------|-----|-----|
|       | Pos   | Neg | Pos   | Neg | Pos   | Neg | Pos    | Neg |     |
| LSTM0 | 60    | 90  | 53    | 91  | 51    | 92  | 53     | 91  | 83  |
| LSTM1 | 67    | 85  | 70    | 81  | 64    | 86  | 36     | 96  | 82  |
| LSTM6 | 56    | 90  | 42    | 93  | 37    | 96  | 50     | 93  | 83  |

**OA*:** Overall accuracy for best performing batch size(BS=16)

Dropout layers are generally used to solve the issue of overfitting. Dropout layers work by randomly assigning a predetermined number of weights of some hidden layers 0 during the training phase. In this case, it is established that better performance can be achieved by having a smaller training dataset, which also reduces the training time and memory which is highly desirable in this field.

TABLE 6: EFFECT OF TRAINING DATASET SIZE

|       | Split=0.33 |     |         | Split=0.4 |     |         |
|-------|------------|-----|---------|-----------|-----|---------|
|       | Pos        | Neg | Overall | Pos       | Neg | Overall |
| LSTM0 | 53         | 91  | 84      | 44        | 94  | 84      |
| LSTM1 | 70         | 81  | 80      | 50        | 93  | 84      |
| LSTM6 | 42         | 93  | 83      | 58        | 88  | 82      |

V. Conclusion

From the above set of experiments done in this work, Sentiment Analysis using LSTMs and its simplified versions was carried out successfully. All the networks predict negative sentiments more accurately than the positive sentiments. This was expected due the unbalanced nature of the training dataset. As per the author's understanding, artificially balancing the training dataset should improve the positive sentiment classification and also the overall performance of the networks. Using a CNN layer and then having a LSTM layer i.e. first extracting the features and then passing it through a LSTM layer is recommended for improved performance. Also, adding more dense layers after the LSTM block doesn't really improve the performance; rather it is increasing the number of trainable parameters and the training time. Hence, adding multiple dense layers after the LSTM block isn't recommended. The learning rate 0.001 or the default learning rate for RMSprop and Adam (Keras) works best for the architectures explored in this work. Either of RMSprop or Adam work good on this dataset but when we consider the positive and negative sentiment classification individually, RMSprop has a slight advantage as it matches or beats the performance of the Adam optimizer. Lastly, the optimum performance is achieved at a batch size equal to 16 and a training-validation data split ratio of 0.4. On the whole, LSTM0 or the standard LSTM and LSTM6 perform the best while LSTM1 doesn't perform well comparatively. It then makes sense to use the slimmed down LSTM6 model as it has comparable performance, and is computationally more efficient. In general, adding the bidirectional LSTM layer wrapper has improved the performance of the system as a whole. The work presented here is conceptual, and is subject to so many parameters and techniques that can improve the overall system or performance, and further insights might be presented in future publications.